\documentclass[sigconf,nonacm]{acmart}
\AtBeginDocument{%
  }

\usepackage{wrapfig}

\usepackage{newfloat}
\usepackage{listings}
\usepackage{tikz}
\usepackage{booktabs}
\usepackage{multirow}
\usepackage{tabularx}
\usepackage{amsmath}
\begin{document}

\title{Boosting Large Language Models for Mental Manipulation Detection via Data Augmentation and Distillation}


\author{Yuansheng Gao}
\email{y.gao@zju.edu.cn}
\affiliation{%
  \institution{Zhejiang University}
  \city{Hangzhou}
  \country{China}
}

\author{Peng Gao}
\email{PengGaoZJU@hotmail.com}
\affiliation{%
  \institution{Zhejiang University}
  \city{Hangzhou}
  \country{China}
}

\author{Han Bao}
\email{baohan21@zju.edu.cn}
\authornotemark[1]
\affiliation{%
  \institution{Zhejiang University}
  \city{Hangzhou}
  \country{China}
}

\author{Bin Li}
\email{b.li2@siat.ac.cn}
\affiliation{%
  \institution{Shenzhen Institute of Advanced Technology, Chinese Academy of Sciences}
  \city{Shenzhen}
  \country{China}
}

\author{Jixiang	Luo}
\email{jixiangluo85@gmail.com}
\affiliation{%
  \institution{Institute of Artificial Intelligence (TeleAI), China Telecom}
  \city{Shanghai}
  \country{China}
}

\author{Zonghui	Wang}
\email{zhwang@zju.edu.cn}
\affiliation{%
  \institution{Zhejiang University}
  \city{Hangzhou}
  \country{China}
  }
\authornote{Corresponding author.}

\author{Wenzhi Chen}
\email{chenwz@zju.edu.cn}
\affiliation{%
  \institution{Zhejiang University}
  \city{Hangzhou}
  \country{China}
}

\renewcommand{\shortauthors}{Yuansheng Gao et al.}

\begin{abstract}

Mental manipulation on social media poses a covert yet serious threat to individuals' psychological well-being and the integrity of online interactions. Detecting such behavior is challenging due to the difficult-to-annotate training data, its highly covert and multi-turn nature, and the lack of real-world datasets. To address these challenges, we propose MentalMAD, a framework that enhances large language models for mental manipulation detection. Our approach consists of three key components: EvoSA, an annotation-free data augmentation method that combines evolutionary operations with speech-act-aware prompting; teacher-model-generated complementary-task supervision; and Complementary-Convergent Distillation, a phase-wise strategy for transferring manipulation-specific knowledge to student models. We then constructed the ReaMent dataset, comprising 5,000 real-world-sourced dialogues. Extensive experiments show that MentalMAD improves accuracy by 14.0\%, macro-F1 by 27.3\%, and weighted F1 by 15.1\% over the strongest baseline. The code and the dataset are publicly available at https://github.com/Yuansheng-Gao/MentalMAD.

\end{abstract}



\maketitle

\section{Introduction}

\begin{figure}[t]
\centering
\includegraphics[width=\linewidth]{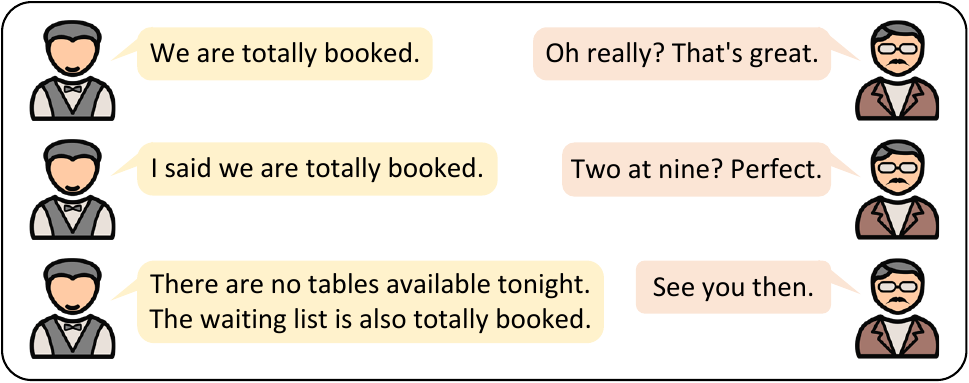}
\caption{A case of mental manipulation.}
\label{fig:task_exp}
\end{figure}

Mental manipulation is a covert form of psychological control conveyed through language and interaction \citep{al2017pragmatic}, as illustrated in Figure~\ref{fig:task_exp}. Building on this characterization, \citet{wang-etal-2024-mentalmanip} further defines mental manipulation as using language to influence, alter, or control an individual's psychological state or perception for the manipulator's benefit. On social media, where information spreads rapidly and remains persistently visible, this phenomenon becomes more widespread and more harmful. Prior research shows that mental manipulation, including gaslighting and coercive persuasion, can cause substantial psychological harm \citep{barnhill2022philosophy, ramiro2019online}. Manipulative cyberbullying that distorts a victim's sense of reality significantly increases suicidal ideation and attempts, with odds ratios of 2.23 and 2.55 \citep{van2014relationship}. Nearly half of U.S. adults report experiencing partner behaviors grounded in mental manipulation, such as gaslighting or blame shifting \citep{creech2023evaluation}. A meta-analysis similarly finds that coercive control, a core form of mental manipulation, is associated with heightened risks of post-traumatic stress symptoms ($r = 0.32$) and depressive symptoms ($r = 0.27$) \citep{lohmann2024trauma}.

Despite the importance of detecting mental manipulation, existing methods still rely mainly on simple heuristics based on large language models (LLMs) \citep{meng2025sanitize, yuan2025reflectdiffu, yuan2024cultural}, which are insufficient for its covert, context-dependent nature. In this context, current research faces three major challenges.
\textbf{Challenge~1: Difficult-to-annotate training data.} Manipulative intent is rarely explicit but instead emerges from subtle cues across multiple turns, making annotation slow and costly. Although LLMs \citep{grattafiori2024llama, yuan2024reversal} can generate synthetic data, such outputs often diverge from human judgment and still require extensive manual verification \citep{wang-etal-2024-mentalmanip}.
\textbf{Challenge~2: The covert and multi-turn nature of mental manipulation hinders detection.} Mental manipulation is difficult to detect in practice because manipulative intent is rarely expressed explicitly. It typically emerges through subtle shifts in framing and gradual pressure across conversational turns \citep{sheshanarayana2025unmasking}. Individual utterances often appear harmless, and harmful intent becomes clear only when the dialogue is considered as a whole. This implicit and multi-turn nature makes automatic detection intrinsically challenging. In contrast, toxicity detection focuses on utterances that are rude, disrespectful, or hateful \citep{zhang2024efficient}. Although both belong to harmful content detection, the conceptual definitions of the two tasks are different. As a result, techniques designed for toxicity detection \citep{meguellati2025llm, vishwamitra2024moderating, kang2024implanting} cannot be directly applied, and existing LLM-based methods \citep{wang-etal-2024-mentalmanip, ma-etal-2025-detecting, yang2024enhanced} still show limited improvement on mental manipulation detection.
\textbf{Challenge~3: Lack of real-world datasets.} MentalManip~\citep{wang-etal-2024-mentalmanip} consists of roughly 4,000 movie-derived dialogues, while LegalCon~\citep{sheshanarayana2025claim} contains 1,038 courtroom exchanges, providing limited coverage of real mental manipulation behavior.

\begin{figure*}[t]
\centering
\includegraphics[width=0.98\linewidth]{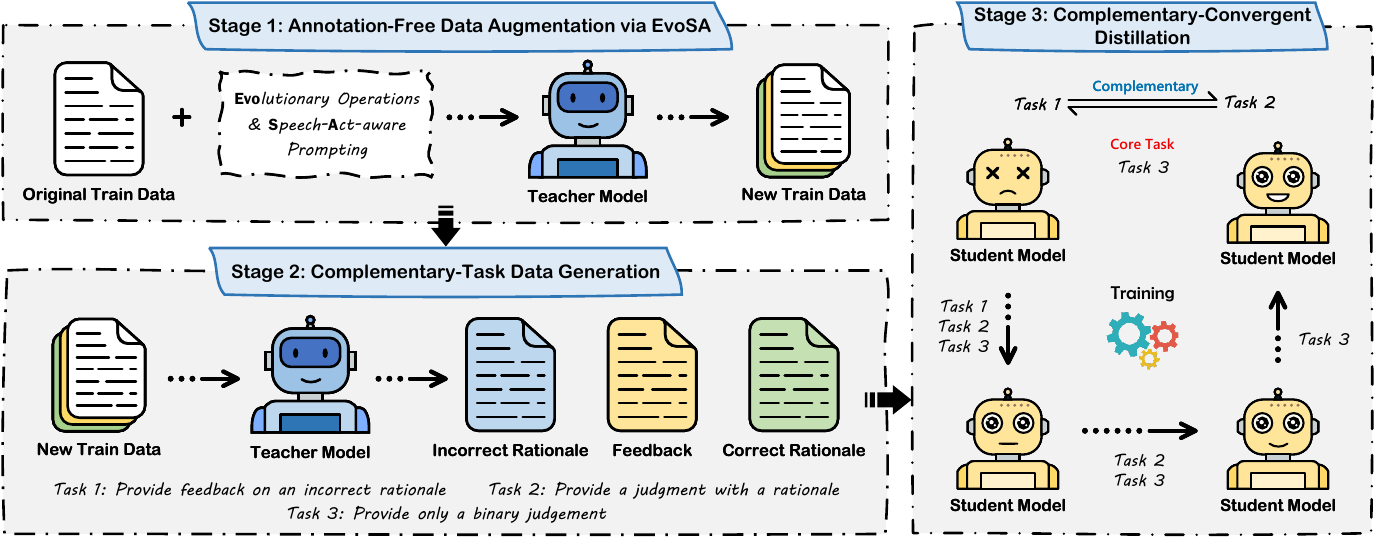
}
\caption{Overall workflow of the proposed MentalMAD.}
\label{fig:workflow}
\end{figure*}

To address these challenges, we propose \textbf{MentalMAD}, a framework for \textbf{Mental} \textbf{M}anipulation detection through data \textbf{A}ugmentation and \textbf{D}istillation (Figure~\ref{fig:workflow}). The first component, \textbf{EvoSA}, tackles the challenge of difficult-to-annotate training data with an annotation-free augmentation strategy. It combines \textbf{Evo}lutionary operations with \textbf{S}peech-\textbf{A}ct-aware prompting to generate label-preserving dialogues that remain coherent and natural. To better handle the covert and multi-turn nature of mental manipulation in detection, we introduce \textbf{Co}mplementary \textbf{Co}nvergent \textbf{Distill}ation (\textbf{CoCoDistill}). Negative instances often lack salient manipulative cues, resulting in weak rationales (Figure~\ref{fig:dss_info}) that hinder direct distillation \citep{huffaker2020crowdsourced}. CoCoDistill uses complementary training data and a staged training scheme that first covers all tasks and then converges to the core objective, enabling the student model to better acquire the essential intent of manipulation. Finally, we construct \textbf{ReaMent}, a dataset of 5,000 dialogues from unscripted human-human interactions in publicly available web videos (YTD-18M \citep{han2023champagne}) to support \textbf{Rea}l-world \textbf{Ment}al manipulation detection. ReaMent compensates for the lack of real-world datasets and provides a realistic, challenging testbed that complements existing benchmarks. Our key contributions are:
\begin{itemize}

\item\textbf{MentalMAD}, a novel framework that integrates EvoSA-based data augmentation to overcome annotation challenges and CoCoDistill with complementary tasks to enhance LLMs' capabilities in detecting mental manipulation.

\item\textbf{ReaMent}, a human-annotated dataset of 5,000 real-world-sourced dialogues that fills the gap left by the absence of real-world data in existing benchmarks.

\item\textbf{Extensive experiments} show that our approach both closes the student-teacher performance gap and outperforms state-of-the-art (SOTA) LLMs on accuracy and F1~scores.

\end{itemize}

\section{Related Works}

\textbf{Mental Manipulation Detection.}
There is a substantial conceptual distinction between mental manipulation and toxic language, which renders existing toxicity detection techniques \citep{meguellati2025llm, zhang2024efficient, vishwamitra2024moderating, kang2024implanting} unsuitable for direct application to mental manipulation detection. Current datasets provide only limited coverage: MentalManip \citep{wang-etal-2024-mentalmanip} is largely scripted or synthetic, while LegalCon \citep{sheshanarayana2025claim} is small, domain-constrained, and contains just over 1,000 courtroom instances. Recent approaches, including chain-of-thought prompting (CoT) \citep{yang2024enhanced}, intent-aware prompting (IAP) \citep{ma-etal-2025-detecting} and CLAIM \citep{sheshanarayana2025unmasking}, have improved detection but still struggle to capture the covert intent, evolving tactics, and pragmatic cues that characterize real-world manipulative behavior. Our work addresses these limitations through ReaMent and a training framework specifically designed for mental manipulation detection.

\textbf{LLM-Based Dialogue Data Augmentation.}
LLMs have become an increasingly common choice for dialogue data augmentation~\citep{dai2025auggpt}. Prior work shows that they can generate diverse dialogue variants, such as AugESC for emotional-support dialogues~\citep{zheng2023augesc}, summary-guided generation for low-resource open-domain dialogues~\citep{liu2024controllable}, and knowledge-driven prompting for multi-turn psychological dialogues~\citep{jiang2024data}. Beyond generation, PromptMix~\citep{sahu2023promptmix} enhances diversity and robustness via prompt-label interpolation during distillation. However, most augmentation approaches emphasize semantic variation while failing to preserve the pragmatic cues essential for manipulation detection; label drift and the loss of discourse-level structure remain major challenges~\citep{ding2024data}. Motivated by these limitations, EvoSA innovatively integrates evolutionary operations with speech-act-aware prompting to produce label-preserving, pragmatically coherent augmented dialogues.

\textbf{Knowledge Distillation for LLMs.}
Recent advances in knowledge distillation \citep{hinton2015distilling} leverage intermediate reasoning signals or multi-task supervision, including PINTO \citep{wang2022toxicity}, Distilling Step-by-Step \citep{hsieh2023distilling}, TinyLLM \citep{tian2025beyond}, and SuperCorrect \citep{yang2025supercorrect}. Distillation has also been explored for harmful language detection \citep{zhang2024efficient}. Although these methods are effective, they assume that both positive and negative instances provide informative rationales. This assumption breaks down in mental manipulation detection, where negative examples often lack clear evidence. In addition, existing methods do not consider how auxiliary training tasks may affect the core task. To address these issues, our CoCoDistill introduces complementary tasks that supply rich information from both positive and negative instances. The model first benefits from jointly learning all tasks and then gradually converges to the core binary judgment task, thereby enhancing its ability to detect mental manipulation.

\section{Methods}

As shown in Figure~\ref{fig:workflow}, MentalMAD consists of three stages. Stage~1 expands the training set with EvoSA. Stage~2 utilizes a teacher model to generate complementary supervision signals, providing rich information from both positive and negative cases. Stage~3 employs CoCoDistill, a phase-wise distillation method, to enhance the model's ability to detect mental manipulation.

\subsection{Data Augmentation via EvoSA}

\textbf{Motivation.} In evolutionary computation, operations such as selection, crossover, and mutation help maintain both the quality and the diversity of a population \citep{YANG2025101838}. In our dialogue corpus, diversity similarly arises from the different speech acts embodied in each dialogue \citep{searle1980speech}. Inspired by this analogy, we aim to incorporate evolutionary operations and speech-act awareness into LLM-based dialogue augmentation. However, while such operations help preserve dialogue quality, they still require manual labeling. To remove this dependence, we instead select dialogues that share the same label and use constrained prompting to encourage the LLM to maintain label consistency during generation.

\textbf{Method.} Building on these ideas, we design EvoSA as a label-preserving augmentation method guided by evolutionary operations and speech-act-aware prompting. Given two parent dialogues with the same label, the teacher model first selects utterances that instantiate distinct speech acts and conversational strategies, then performs recombination and stronger content mutations to construct an initial child dialogue (Steps 1-3). The child dialogue is then refined in a separate step, where the model reasons about why the parent dialogues carry their label and optimizes the child dialogue once to better align its pragmatic cues with the target label while keeping it coherent and natural (Steps 4–7). The complete seven-step prompt template is shown in Figure~\ref{fig:evosa_prompt}.

Formally, given a teacher model $\mathcal{T}$ and a dataset $\mathcal{D} = \big\{(x_i, y_i)\big\}_{i=1}^n$, EvoSA samples two parent dialogues with identical labels and generates a new dialogue as
\begin{equation}
(x_k', y_k') = \mathcal{T}\big((x_i, y_i), (x_j, y_j), p_{EvoSA}\big) \quad\text{s.t.} \quad i \neq j, \; y_i = y_j,
\end{equation}
where $p_{EvoSA}$ represents EvoSA prompts. Let $\mathcal{D}$ contains $n_+$ positive and $n_-$ negative dialogues. The expanded dataset becomes

\begin{equation}
\mathcal{D'} = \mathcal{D} \cup \big\{(x_i'^{+}, y_i'^{+})\big\}_{i=1}^{n_+'} \cup \big\{(x_j'^{-}, y_j'^{-})\big\}_{j=1}^{n_-'},
\label{eq:2}
\end{equation}
where $(x_i'^{+}, y_i'^{+})$ and $(x_j'^{-}, y_j'^{-})$ denote the synthesized positive and negative samples, and $n_+'$ and $n_-'$ are the corresponding numbers of new instances. The numbers of newly added samples satisfy
\begin{equation}
n_+' \in \big[1, C (n_+, 2)\big] \cap \mathbb{Z} \quad
n_-' \in \big[1, C (n_-, 2)\big] \cap \mathbb{Z}.
\end{equation}


\subsection{Complementary-Task Data Generation}

\begin{figure}
\centering
\includegraphics[width=0.92\linewidth]{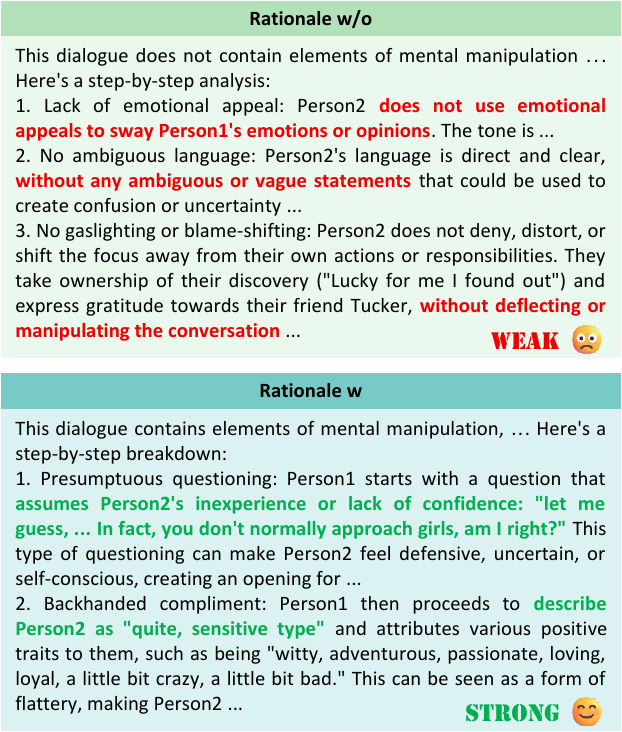}
\caption{Example rationales for a dialogue with ("Rationale w") and without ("Rationale w/o") mental manipulation.}
\label{fig:dss_info}
\end{figure}

\textbf{Motivation.} \citet{huffaker2020crowdsourced} note that in emotionally manipulative detection, positive instances often contain coercive or suggestive expressions, making them easier to identify and justify. Negative instances lack such cues, which makes non-manipulative judgments harder to explain. As shown in Figure~\ref{fig:dss_info}, this asymmetry also appears in our task: when generating rationales for negative instances, LLMs often struggle to produce sufficiently discriminative content, reducing the effectiveness of distillation.

\textbf{Method.} To mitigate this asymmetry, we introduce three tasks. Task 1 and Task 2 are complementary, and together they enrich the supervision signal for Task 3, which is the primary objective. Specifically, we define the tasks as follows:
\begin{itemize}

\item \textbf{Task 1:} Provide feedback on an incorrect rationale;
\item \textbf{Task 2:} Provide a judgment with a rationale;
\item \textbf{Task 3:} Provide only a binary judgment.

\end{itemize}

According to our analysis, even in non-manipulative dialogues, the model's faulty rationales often contain elaborate but flawed reasoning. This makes the feedback on such rationales rich and informative. As a result, Task~1 strategically compensates for the limited quality of rationales in Task~2, offering the student model more discriminative and instructive supervision signals.

For Task 1, an incorrect rationale $r_i^{-}$ is produced through
\begin{equation}
r_i^{-} = \mathcal{T}\big(P_{r-}(x_i',y_i')\big),
\end{equation}
where $P_{r-}$ denotes the prompt for producing incorrect rationale. Feedback on this rationale $f_i$ is then generated as

\begin{equation}
f_i = \mathcal{T}\big(P_{f}(x_i',y_i', r_i^{-})\big),
\end{equation}
where $P_{f}$ being the prompt for rationale feedback.

For Task 2, the correct rationale $r_i^{+}$ is obtained using:
\begin{equation}
r_i^{+} = \mathcal{T}\big(P_{r+}(x_i',y_i')\big),
\end{equation}
where $P_{r+}$ drives the teacher model to explain the correct label.

Task 3 relies solely on the binary label and does not require additional generation.
By supplying complementary signals for both positive and negative instances, this design compensates for the weak rationales often present in negative examples and provides the student model with richer supervision for distillation. All task prompts are listed in Appendix~\ref{sec:prompt}.

\subsection{Complementary-Convergent Distillation}

\textbf{Motivation.} Learning from multiple reasoning signals can strengthen mental manipulation detection, but treating these signals as parallel tasks often causes gradient conflicts. Since the ultimate goal is accurate binary classification, the training schedule must integrate complementary reasoning sources in a way that enhances supervision without disrupting optimization toward the core objective.

\textbf{Method.} CoCoDistill uses the augmented dataset $\mathcal{D'}$ and distills teacher generated signals into a student model $\mathcal{S}$. To benefit from the complementary tasks while minimizing their impact on the binary classification task, the distillation process is organized into three phases. Phase~1 trains on all tasks, Phase~2 uses Tasks~2 and~3, and Phase~3 focuses solely on Task~3.

\textbf{Task 1.} The student receives an incorrect rationale without true labels and produces feedback
\begin{equation}
f_i^{\mathcal{S}} \;=\; \mathcal{S}\!\big(P_f^{\mathcal{S}}(x_i',\,y_i',\,r_i^-)\big),
\end{equation}
where $P_f^{\mathcal{S}}$ denotes the feedback prompt provided to the student. The corresponding loss is
\begin{equation}
\mathcal{L}_1 \;=\; \frac{1}{|\mathcal{D'}|} \sum_{i=1}^{|\mathcal{D'}|} \ell\big(f_i^{\mathcal{S}},\,f_i\big),
\end{equation}
where $\ell(\cdot,\cdot)$ denotes the loss of cross-entropy. This objective allows the student model to learn a rich reverse-reasoning signal, preventing the adverse effects caused by the imbalance of reasoning information across categories.

\textbf{Task 2.} The student generates a label and a supporting rationale
\begin{equation}
r_i^{\mathcal{S}} \;=\; \mathcal{S}\big(P_r^{\mathcal{S}}(x_i',\,y_i')\big),
\end{equation}
where $P_r^{\mathcal{S}}$ is the prompt for Task~2. The corresponding loss is
\begin{equation}
\mathcal{L}_2 \;=\; \frac{1}{n'} \sum_{i=1}^{n'} \ell\big(r_i^{\mathcal{S}},\,r_i^+\big).
\end{equation}
This design enables the student model to learn high-quality rationales, improving binary classification accuracy.

\textbf{Task 3.} The student outputs a binary judgment
\begin{equation}
y_i^{\mathcal{S}} \;=\; \mathcal{S}\big(P_y^{\mathcal{S}}(x_i')\big),
\end{equation}
where $P_{y}^{\mathcal{S}}$ prompts the binary judgment.
The loss for Task~3 is
\begin{equation}
\mathcal{L}_3 \;=\; \frac{1}{n'} \sum_{i=1}^{n'} \ell\big(y_i^{\mathcal{S}},\,y_i'\big).
\end{equation}
This objective provides direct supervision on the final decision, ensuring that the learned reasoning ultimately supports accurate binary classification, which is the core goal of the model.

\textbf{Phase-Wise Objectives.} The objectives for the three phases are
\begin{align}
\text{Phase 1:}\quad & \mathcal{L}_1^{\mathcal{S}} \;=\; \mathcal{L}_1 + \mathcal{L}_2 + \mathcal{L}_3;\\
\text{Phase 2:}\quad & \mathcal{L}_2^{\mathcal{S}} \;=\; \mathcal{L}_2 + \mathcal{L}_3;\\
\text{Phase 3:}\quad & \mathcal{L}_3^{\mathcal{S}} \;=\; \mathcal{L}_3.
\end{align}
This sequence allows the student model to first acquire broad task knowledge and then gradually concentrate on the core binary objective. The final phase ensures that the model fully consolidates the representations required for detecting mental manipulation.

\section{ReaMent Dataset}
To fill the gap of real-world data in this field, we constructed ReaMent. This section describes its creation process, including data
collection, annotation, statistics, and annotation-quality analysis.

\subsection{Data Collection}

We construct ReaMent on top of YTD-18M \citep{han2023champagne}, a corpus of over 18 million dialogue-like segments reconstructed from automatically transcribed, unscripted interactions in real-world web videos. The dataset spans a wide range of everyday scenarios, including third-person social exchanges as well as first-person situational interactions. These varied settings provide rich pragmatic contexts for studying mental manipulation. We extract the textual transcripts and randomly sample 102{,}000 dialogues as the candidate pool. Since YTD-18M is manually curated, only extremely short or malformed dialogues are removed. To protect anonymity and reduce potential bias, all speaker names are replaced with "Person1" and "Person2".

Because instances of mental manipulation are rare in natural dialogue, we apply a lightweight pre-filter to retrieval to reduce annotation cost. Using LLaMA3-70B-Instruct~\citep{grattafiori2024llama} and mental manipulation-related key phrases from \citet{wang-etal-2024-mentalmanip} (see Appendix~\ref{app:keyphrases}), we first detect dialogues that may contain manipulative content. To identify key-phrase matches, we use a length-adaptive criterion: a dialogue is flagged if any sentence shares at least $P\%$ of its tokens with a key phrase, with $P$ shown in Table~\ref{tab:keyphrase_match}. LLM outputs are used only for candidate retrieval, not labeling. We then take all pre-filtered dialogues and additionally randomly sample dialogues from the unflagged pool, resulting in 9{,}401 candidate dialogues for annotation.

\subsection{Data Annotation}
\subsubsection{Annotator Recruitment and Training}
Annotators are recruited through a voluntary application process from undergraduate and graduate students who are native or fluent English speakers. All candidates receive dedicated training on the task. After completing the training, they take a qualification test of 100 dialogues sampled from MentalManip\textsubscript{con}, and only those achieving at least 85\% accuracy are retained. This process yields 12 qualified annotators with diverse genders, academic backgrounds (e.g., mechanics, computer science, psychology), and cultural backgrounds (e.g., born in Malaysia, China, and Australia).

\subsubsection{Annotation Procedure}
To distribute workload and reduce annotator fatigue, we divide the 12 annotators into four independent groups of three. Each dialogue is labeled by exactly three annotators, with disjoint subsets assigned to each group. Annotators are compensated according to local norms for similar tasks and are
given the following guideline:

{\small
\noindent\texttt{\textquotesingle\textquotesingle\textquotesingle\\
Based on the definition of mental manipulation, please determine whether the given dialogue contains elements of mental manipulation. If it does, label it as 1; otherwise, label it as 0. In addition, provide your confidence level in the annotation on a 5-point scale (1 = very uncertain, 5 = very confident).\\
\\
\#\#\# Definition:\\
Mental manipulation is using language to influence, alter, or control an individual's psychological state or perception for the manipulator's benefit.\\
\\
\#\#\# Dialogue:\\
<insert dialogue>\\
\textquotesingle\textquotesingle\textquotesingle}}

\subsection{Dataset Statistics}

After annotation, we filter out low-quality labels and construct the ReaMent dataset with 5{,}000 high-quality instances. Dialogues with full annotator agreement form a strict subset, ReaMent\textsubscript{con}. Table~\ref{tab:stats} reports dataset sizes and basic statistics. Compared with LegalCon (1{,}038 instances), which centers on courtroom interactions, and MentalManip (4{,}000 instances), which is derived from scripted movie dialogues, ReaMent offers a larger scale and draws from more authentic and diverse real-world sources. Collected from large-scale web videos, ReaMent spans a much broader range of conversational scenarios. These include third-person social exchanges such as interviews and group discussions, as well as everyday situational interactions such as outdoor activities and instructional conversations. This diversity yields more natural and spontaneous interaction patterns than those found in scripted movie language. Figures~\ref{fig:case_no} and~\ref{fig:case_yes} present representative examples. Dialogues in ReaMent better capture everyday interpersonal communication, including hesitations, interruptions, and incomplete expressions, whereas MentalManip often uses dramatized or stylized expressions.

\subsection{Analysis of Annotation Quality}
To assess annotation reliability, we compute Fleiss'~$\kappa$ over the full dataset, obtaining $\kappa = 0.52$, which indicates moderate agreement. This level of consistency aligns with expectations, as the judgment of manipulation is inherently subjective.

Although the overall consensus is moderate, this does not imply low annotator certainty. In subjective tasks like manipulation detection, annotators may be confident in their own judgments even when they disagree. To capture this aspect of reliability, we compute a confidence-weighted score for each instance:
\begin{equation}
v_k = \frac{\sum_{i=1}^{n_a} c_{ik} \cdot I(a_{ik}, t_k)}{\sum_{i=1}^{n_a} c_{ik}},
\end{equation}
where $n_a$ is the number of annotators, $c_{ik}$ denotes the annotator's confidence (1–5), and $t_k$ is the majority-vote label. The scores yield a mean of 0.87, a first quartile of 0.64, and a median of 1.00, suggesting that despite the task's inherent subjectivity, most instances achieve high-confidence agreement among annotators.

\begin{table}
\small
\centering
\begin{tabular}{c|cccc}
\toprule
\textbf{Key Phrase Length} & $\le 4$ & $\le 6$ & $\le 10$ & $> 10$ \\
\midrule
\textbf{Matching Percentage $P$} & 100\% & 90\% & 80\% & 70\% \\
\bottomrule
\end{tabular}
\caption{Length-adaptive matching criterion.}
\label{tab:keyphrase_match}
\end{table}

\begin{table}
\small
\centering
\begin{tabular}{lcc}
\toprule
\textbf{Variant} & \textbf{ReaMent\textsubscript{con}} & \textbf{ReaMent} \\
\midrule
Sample Size & 3{,}355 & 5{,}000 \\
Avg. Turns & 4.0 & 3.8 \\
Avg. Words & 81.2 & 80.2 \\
No. of Yes Labels & 2{,}362 (70.4\%) & 3{,}417 (68.3\%) \\
No. of No Labels & 993 (29.6\%) & 1{,}583 (31.7\%) \\
\bottomrule
\end{tabular}
\caption{Basic statistics of our ReaMent dataset.}
\label{tab:stats}
\end{table}

\section{Experiments}

This section evaluates MentalMAD through baselines and ablations, examines its scalability, assesses EvoSA, and presents case studies to illuminate the sources of MentalMAD's strong performance.

\begin{table*}[t]
\small
\centering
\begin{tabular}{llccccc|ccccc|ccccc}
\toprule
\multirow{2}{*}{\textbf{Model}} & \multirow{2}{*}{\textbf{Method}} 
& \multicolumn{5}{c|}{\textbf{MentalManip (\%)}} 
& \multicolumn{5}{c|}{\textbf{ReaMent (\%)}} 
& \multicolumn{5}{c}{\textbf{LegalCon (\%)}} \\
\cmidrule(lr){3-7} \cmidrule(lr){8-12} \cmidrule(lr){13-17}
& & Acc & Pre & Re & F1\textsubscript{m} & F1\textsubscript{w} 
  & Acc & Pre & Re & F1\textsubscript{m} & F1\textsubscript{w}
  & Acc & Pre & Re & F1\textsubscript{m} & F1\textsubscript{w} \\
\midrule
\multicolumn{2}{l}{DeepSeek-R1}     & 73.9 & 73.0 & 98.8 & 57.1 & 67.4 &70.9 &72.6 &94.3 &52.8 &64.8  &62.0 &94.4 &40.2 &61.4 &60.3 \\
\multicolumn{2}{l}{GPT-5 Chat}        &72.6 &72.9 &96.0 &57.0 &66.9 &70.9 &72.6 &94.5 &52.5 &64.6 &64.9 &93.5 &45.7 &64.6 &63.9 \\
\multicolumn{2}{l}{Claude-Haiku 4.5}       &74.6 &76.2 &92.1 &64.9 &71.9 &62.7 &74.8 &71.0 &56.7 &63.3 &63.0 &98.1 &40.2 &62.2 &61.1 \\
\multicolumn{2}{l}{LLaMA3-70B-Instruct}       & 72.0 & 76.1 & 86.8 & 63.7 & 70.3 &70.8 &72.5 &94.1 &53.0 &64.8  &50.5 &90.0 &21.3 &47.3 &44.5 \\
\midrule
\multirow{7}{*}{Qwen2.5-3B-Instruct} 
  & Vanilla  & 62.4 & 79.3 & 61.8 & \underline{60.3} & 63.8 &47.8 &74.4 &39.4 &47.5 &49.2  &50.0 &59.5 &56.7 &48.1 &50.3 \\
  & CoT      & 39.6 & \textbf{85.9} & 15.1 & 37.4 & 33.0 &33.8 &\textbf{86.8} &7.0 &29.8 &22.9  &42.3 &59.0 &18.1 &39.9 &37.2 \\
  & IAP      & 46.1 & \underline{83.5} & 27.5 & 45.8 & 44.1 &35.6 &\underline{85.7} &10.2 &32.6 &26.7  &43.3 &69.6 &12.6 &38.5 &34.7 \\
  & SP       & 51.7 & 75.3 & 44.8 & 51.2 & 53.1 &50.9 &79.8 &40.3 &50.7 &51.9  &43.3 &\textbf{100.0} &7.1 &35.5 &30.6 \\
  & SFT      & 73.2 & 73.3 & \textbf{96.3} & 58.3 & 67.8  &\underline{72.6} &73.8 &94.5 &\underline{56.9} &\underline{67.5}  &\underline{68.3} &71.0 &81.1 &\underline{65.0} &\underline{67.3} \\
  & DSS  & \underline{73.8} & 73.9 & \underline{95.8} & 60.0 & \underline{69.0}  &70.3 &70.7 &\textbf{98.7} &44.0 &59.7  &65.9 &67.9 &\underline{83.5} &60.8 &63.9 \\
  & Ours     & \textbf{75.6} &79.6 & 87.1 & \textbf{69.5} & \textbf{74.8} &\textbf{78.2} &78.5 &\underline{95.1} &\textbf{68.5} &\textbf{75.6}  &\textbf{74.5} &\underline{75.3} &\textbf{86.6} &\textbf{71.8} &\textbf{73.7} \\
\midrule
\multirow{7}{*}{Phi-3.5-Mini-Instruct} 
  & Vanilla  & 70.0 & 71.6 & \underline{93.8} & 53.4 & 64.0  &71.7 &72.9 &\underline{95.1} &53.9 &65.5  &55.8 &62.2 &70.1 &51.5 &54.7 \\
  & CoT      &70.7 & 71.9 & \textbf{94.5} & 54.1 & 64.7  &72.7 &73.3 &\textbf{96.2} &55.2 &66.6  &60.1 &67.7 &66.1 &58.3 &60.2 \\
  & IAP      & 66.4 & 77.3 & 72.7 & 61.9 & 66.9  &68.3 &79.6 &73.7 &63.7 &68.9  &56.2 &70.5 &48.8 &56.2 &56.5 \\
  & SP       & 66.6 & 71.9 & 84.6 & 55.1 & 63.8 &69.9 &73.4 &89.6 &56.0 &66.1  &61.5 &\underline{96.1} &38.6 &60.7 &59.5 \\
  & SFT      & 74.1 & \underline{78.8} & 85.6 & \underline{67.8} & 73.2  &74.5 &76.7 &91.8 &63.6 &71.7  &59.1 &68.4 &61.4 &58.1 &59.6 \\
  & DSS  & \underline{75.0} & 77.2 & 90.6 & 66.5 & \underline{72.9}  &\underline{79.1} &\underline{82.0} &90.0 &\underline{73.0} &\underline{78.3}  &\underline{75.5} &86.5 &\underline{70.9} &\underline{75.2} &\underline{75.8} \\
  & Ours     & \textbf{76.8}	&\textbf{81.3}	&86.4	&\textbf{71.7}	&\textbf{76.3}	&\textbf{81.2}	&\textbf{89.5}	&83.1	&\textbf{78.5}	&\textbf{81.6}	&\textbf{86.1}	&\textbf{97.1}	&\textbf{79.5}	&\textbf{85.9}	&\textbf{86.2} \\
\midrule
\multirow{7}{*}{MiniCPM3-4B} 
  & Vanilla  & 69.8 & 69.7 & \textbf{99.5} & 44.2 & 58.7  &70.6 &70.7 &\underline{99.4} &43.7 &59.6  &59.6 &60.5 &97.6 &37.3 &45.6 \\
  & CoT      &71.0 &70.7 &99.3 &48.4 &61.5  &70.3 &70.6 &99.2 &43.2 &59.2  &58.2 &59.9 &95.3 &36.8 &44.9 \\
  & IAP      & 69.0 &69.1 &\textbf{99.5} &41.3 &56.7  &70.6 &70.6 &\textbf{99.8} &42.8 &59.1  &60.6 &60.9 &\textbf{99.2} &37.3 &46.1 \\
  & SP       & 71.7 & 71.7 & 97.5 & 53.0 & 64.3 &71.5 &71.6 &98.5 &48.3 &62.4  &73.6 &70.9 &96.1 &67.3 &70.5 \\
  & SFT      & 75.6 &\underline{76.3} &95.8 &\underline{65.8} &\underline{73.1}  &71.8 &71.8 &98.7 &48.8 &62.8  &73.7 &69.9 &\textbf{99.2} &66.5 &69.7 \\
  & DSS  & \underline{77.2} &75.8 &98.5 &65.0 &72.9  &\underline{74.5} &\underline{74.5} &97.0 &\underline{58.6} &\underline{69.0}  &\underline{89.9} &\textbf{94.2} &89.0 &\underline{89.5} &\underline{90.0} \\
  & Ours     & \textbf{78.6} & \textbf{81.2} & 89.8 & \textbf{72.9} & \textbf{77.6} &\textbf{80.0} &\textbf{83.1} &89.8 &\textbf{74.6} &\textbf{79.4}  &\textbf{91.3} &\underline{88.1} &\textbf{99.2} &\textbf{90.5} &\textbf{91.1} \\
\bottomrule
\end{tabular}
\caption{Comparison results. The best and second-best student model results are \textbf{bolded} and \underline{underlined}, respectively.}
\label{tab:main-res}
\end{table*}

\subsection{Experiment Setting}
\textbf{Dataset and Models.} We evaluate on MentalManip\textsubscript{con}, ReaMent\textsubscript{con}, and LegalCon, ensuring reliable labels. All datasets are randomly split into training, validation, and test sets using a 6:2:2 ratio. We use LLaMA3-70B-Instruct as the teacher model and adopt Qwen2.5-Instruct (0.5B, 1.5B, and 3B)~\citep{qwen2.5}, Phi-3.5-Mini-Instruct~\citep{abdin2024phi3}, and MiniCPM3-4B~\citep{hu2024minicpm} as student models.

\textbf{Baselines.} We evaluate five categories: (1) SOTA LLMs in zero-shot setting, including DeepSeek-R1 \citep{guo2025deepseek}, GPT-5 Chat \citep{openai2025gpt5systemcard}, Claude-Haiku 4.5 \citep{anthropic2025haiku45}, and LLaMA3-70B-Instruct; (2) vanilla student models; (3) existing techniques for enhancing LLMs on this task, including CoT, IAP, and SP; (4) a general-purpose distillation method, namely distilling step-by-step (DSS); and (5) SFT \citep{yang2024enhanced}. To highlight the advantage of EvoSA, we perform SFT under four data settings and compare their performance: the original data (Original), duplication-based oversampling (Over), LLM-based label-preserving augmentation (Label-Pre), and our EvoSA-based augmentation (EvoSA).

\textbf{Implementation Details.} To ensure fairness, all baselines follow their original hyperparameter settings, and all augmentation methods use the same data volume and configurations. Experiments run on two NVIDIA H800 GPUs using the Lion optimizer \citep{NEURIPS2023_9a39b492} and LoRA \citep{hu2022lora}. The student model is trained for one epoch per phase with a maximum sequence length of 1,500 tokens, totaling 3 epochs with a batch size of 4 and gradient accumulation of 4. During inference, we restrict the output to 1 token and disable sampling (\texttt{do\_sample=False}). The teacher model uses the same decoding setup with temperature 0 and a maximum output length of 1,024 tokens. Additional details are provided in Appendix~\ref{app:parameters}.

\subsection{Main Results Analysis}

\subsubsection{Baseline Comparison}
The comparison results are shown in Table~\ref{tab:main-res}. While our method does not achieve the highest precision and recall simultaneously, it outperforms others on several key metrics. Although some baselines obtain higher precision or recall, this is mainly due to predicting nearly all instances as \textit{Yes} or \textit{No}. Such degenerate prediction behavior leads to substantially lower accuracy and F1 scores, making them impractical for real-world use. For example, MiniCPM3-4B with IAP attains recall above 0.99 on both datasets, yet its precision remains unacceptably low, suggesting that it predicts nearly all instances as \textit{Yes} and consequently suffers notable drops in accuracy and F1 compared with our method.
Importantly, our approach enables smaller models to outperform large-scale LLMs such as GPT-5 Chat and LLaMA3-70B-Instruct on critical metrics. As shown in Table~\ref{tab:imp_conf}, MentalMAD consistently surpasses the strongest baselines, with the most substantial improvements observed in F1. The confusion matrices further reveal that our method achieves a more balanced FP/FN trade-off, which directly contributes to these gains. These results collectively demonstrate the effectiveness of the proposed MentalMAD.

\begin{table}
\centering
\small
\begin{tabular}{lccc|cccc}
\toprule
\multirow{2}{*}{\textbf{Dataset / Model}}
& \multicolumn{3}{c|}{\textbf{Improvement (\%)}}
& \multicolumn{4}{c}{\textbf{Confusion Matrix}} \\
\cmidrule(lr){2-4} \cmidrule(lr){5-8}
& Acc & F1\textsubscript{m} & F1\textsubscript{w}
& TN & FP & FN & TP \\
\midrule
\multicolumn{8}{l}{\textbf{MentalManip}} \\
Qwen2.5-3B-Instruct & 2.4 & 15.3 & 8.4 & 90 & 90 & 52 & 351 \\
Phi-3.5-mini-Instruct & 2.4 & 5.8 & 4.7 & 100 & 80 & 55 & 348 \\
MiniCPM3-4B & 1.8 & 10.8 & 6.2 & 96 & 84 & 41 & 362 \\
\midrule
\multicolumn{8}{l}{\textbf{ReaMent}} \\
Qwen2.5-3B-Instruct & 7.7 & 20.4 & 12.0 & 76 & 123 & 23 & 449 \\
Phi-3.5-mini-Instruct & 2.6 & 7.5 & 4.2 & 153 & 46 & 80 & 392 \\
MiniCPM3-4B & 7.4 & 27.3 & 15.1 & 113 & 86 & 48 & 424 \\
\midrule
\multicolumn{8}{l}{\textbf{LegalCon}} \\
Qwen2.5-3B-Instruct & 9.1 & 10.5 & 9.5 & 45 & 36 & 17 & 110 \\
Phi-3.5-mini-Instruct & 14.0 & 14.2 & 13.7 & 78 & 3 & 26 & 101 \\
MiniCPM3-4B & 1.6 & 1.1 & 1.2 & 64 & 17 & 1 & 126 \\
\bottomrule
\end{tabular}
\caption{Relative improvement over the best student baseline and confusion matrix (ordered as TN, FP, FN, TP).}
\label{tab:imp_conf}
\end{table}

\subsubsection{Scalability Analysis}
We further assess the scalability of our approach by applying it to smaller models, specifically Qwen2.5-1.5B-Instruct and Qwen2.5-0.5B-Instruct, on the MentalManip dataset. As shown in Table~\ref{tab:size-reament}, although reducing the model size leads to some performance degradation, the 0.5B variant remains competitive with large-scale LLMs such as GPT-5 when enhanced by our framework. These results underscore the ability of our framework to deliver competitive performance even with small-scale models.

\begin{table}
\centering
\small
\begin{tabularx}{\linewidth}{l l *{5}{>{\centering\arraybackslash}X}}
\toprule
Size & Variant & Acc & Pre & Re & F1$_\text{m}$ & F1$_\text{w}$ \\
\midrule
\multirow{3}{*}{3B} 
& Vanilla      & 0.624 & 0.793 & 0.618 & 0.603 & 0.638  \\
& Ours         & 0.768 & 0.795 & 0.896 & 0.703 & 0.756 \\
& $\Delta$ (\%) &+23.1 &+0.3 &+45.0 &+16.6 &+18.5 \\
\midrule
\multirow{3}{*}{1.5B} 
& Vanilla      & 0.415 & 0.844 & 0.189 & 0.401 & 0.365 \\
& Ours         & 0.756 & 0.809 & 0.849 & 0.705 & 0.752  \\
& $\Delta$ (\%) &+82.2	&-4.1	&+349.2	&+75.8	&+106.0 \\
\midrule
\multirow{3}{*}{0.5B} 
& Vanilla      & 0.470 & 0.665 & 0.469 & 0.453 & 0.490 \\
& Ours         & 0.732 & 0.796 & 0.824 & 0.679 & 0.729 \\
& $\Delta$ (\%) & +55.7	& +19.7	& +75.7	& +49.9	& +48.8\\
\bottomrule
\end{tabularx}
\caption{Performance of models across different sizes. $\Delta$~(\%) denotes relative improvement over vanilla baselines.}
\label{tab:size-reament}
\end{table}

\begin{table}
\small
\centering
\begin{tabular}{lccccc}
\toprule
Variant & Acc & Pre & Re & F1\textsubscript{m} & F1\textsubscript{w}\\
\midrule
  w/o EvoSA  &0.760 &0.757 &\textbf{0.970} &0.621 &0.715 \\
  w/o Task 1 &0.769 &0.782 &0.932 &0.673 &0.745 \\
  w/o Task 2 &\underline{0.778} &\underline{0.814} &0.888 &\underline{0.715} &\underline{0.769} \\
  w Joint    &0.763 &0.774 &\underline{0.937} &0.657 &0.735 \\
  w Reverse    &0.751 &0.752 &\underline{0.964} &0.607 &0.704 \\
  Ours     &\textbf{0.797} &\textbf{0.824} &0.905 &\textbf{0.738} &\textbf{0.789} \\
\bottomrule
\end{tabular}
\caption{Ablation results. The best and second-best results are \textbf{bolded} and \underline{underlined}, respectively.}
\label{tab:ablation-res}
\end{table}

\subsection{Ablation Study}
We employ MiniCPM3-4B to evaluate the contributions of each component on the ReaMent dataset, with the corresponding results presented in Table~\ref{tab:ablation-res}. "w Joint" denotes standard joint learning. "w Reverse" indicates the reverse of our distillation order.

After removing EvoSA, all metrics declined except for recall. Although recall reached its highest value in this setting, precision dropped to approximately 0.75, suggesting that the model became biased toward positive predictions, which is suboptimal. Similar patterns were observed under "joint" and "reverse", indicating that the order of task learning plays a critical role in model performance.

Further analysis shows that excluding Task~1 resulted in a marked shift toward positive predictions. While recall improved, the decline in precision caused decreases in both accuracy and F1 score due to the resulting class imbalance. In contrast, the removal of Task~2 led to consistent declines across all evaluation metrics. These findings confirm the importance of both Task~1 and Task~2.

\subsection{Evaluation of EvoSA-Generated Dialogues}

\begin{table}
\centering
\small
\begin{minipage}{0.25\textwidth}
\centering
\begin{tabular}{lccc}
\toprule
Setting        & Yes     & No      & Mean \\
\midrule
E1      & 4.357   & 3.794   & 4.070 \\
E2      & 4.337   & 3.922   & 4.125 \\
E3      & 4.102   & 3.647   & 3.870 \\
Mean    & 4.265   & 3.788   & 4.022 \\
\bottomrule
\end{tabular}
\caption{Results of the quality assessment for dialogues generated by EvoSA.}
\label{tab:annotator-quality}
\end{minipage}
\hfill
\begin{minipage}{0.2\textwidth}
\centering
\begin{tabular}{lccc}
\toprule
Setting & Yes & No  & Total \\
\midrule
$<2$    & 3    & 11  & 14 \\
$\ge2$  & 95   & 91  & 186 \\
Total   & 98   & 102 & 200 \\
\bottomrule
\end{tabular}
\caption{Results of the label consistency assessment for dialogues generated by EvoSA.}
\label{tab:annotator-label}
\end{minipage}
\end{table}

\begin{table}
\small
\centering
\begin{tabular}{lccccc}
\toprule
Variant & Acc & Pre & Re & F1\textsubscript{m} & F1\textsubscript{w}\\
\midrule
  Original &0.591 &0.684 &0.614 &0.581 &0.596 \\
  Over &0.635 &0.711 &\underline{0.677} &0.621 &0.637 \\
  Label-Pre &\underline{0.643} &\underline{0.730} &0.661 &\underline{0.632} &\underline{0.646} \\
  EvoSA    &\textbf{0.668} &\textbf{0.750} &\textbf{0.685} &\textbf{0.659} &\textbf{0.671} \\
\bottomrule
\end{tabular}
\caption{Evaluation of EvoSA's contribution. The best and second-best results are \textbf{bolded} and \underline{underlined}, respectively.}
\label{tab:evosa_contribution}
\end{table}

We recruited three volunteers from the pool of twelve annotators to evaluate the quality of EvoSA-generated child dialogues and assess their label consistency with the corresponding parent dialogues. Each evaluator was compensated at a rate aligned with local compensation norms for similar evaluation tasks. We randomly sampled 200 generated dialogues and asked three evaluators (E1, E2, and E3) to rate them on a 5-point scale, where higher scores indicate better quality. As shown in Table~\ref{tab:annotator-quality}, although dialogues labeled \textit{No} received slightly lower scores than those labeled \textit{Yes}, the overall quality remained high, with an average score of 4.022.

The evaluators also assessed label consistency between each child dialogue and its parent. They marked a child dialogue as 1 if its label matched the parent's, otherwise 0. As shown in Table~\ref{tab:annotator-label}, 186 child dialogues were judged consistent by at least two evaluators ($\geq2$), indicating that EvoSA effectively preserves label consistency.

We next examine whether EvoSA-generated dialogues actually improve model performance. As shown in Table~\ref{tab:evosa_contribution}, all augmented variants use the same amount of training data, yet EvoSA clearly outperforms both oversampling and label-preserving augmentation, indicating that its design yields genuinely more informative training examples rather than merely increasing data volume.

Overall, these results show that EvoSA produces high-quality, label-consistent child dialogues and improves model performance beyond what can be achieved by simply increasing data. Examples and the full evaluation procedure are provided in Appendix~\ref{sec:evosa}.

\subsection{Case Study}

Considering the strong competitiveness of DSS, we analyze representative rationale examples produced by Phi-3.5-Mini-Instruct trained with DSS and our proposed MentalMAD.

\begin{figure}
\centering
\includegraphics[width=0.96\linewidth]{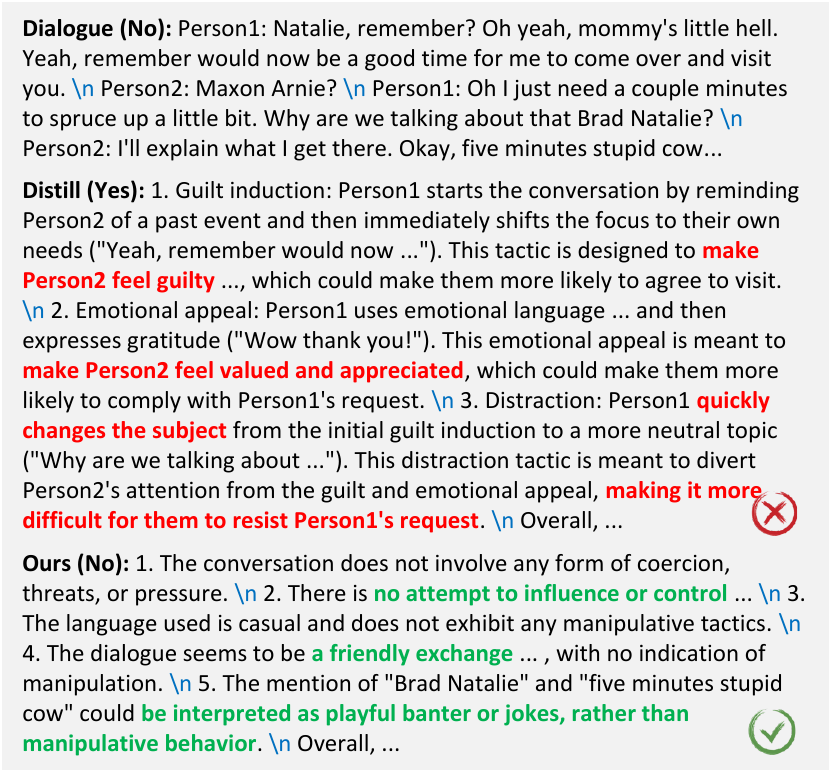}
\caption{Example rationales from the ReaMent dataset.}
\label{fig:case_no}
\end{figure}

In the Figure~\ref{fig:case_no}, DSS incorrectly labels a non-manipulative exchange as manipulative by over-weighting superficial cues such as abrupt topic changes or emotionally colored expressions. These features do not constitute evidence of an attempt to shape the interlocutor's psychological state. In contrast, our method correctly identifies the absence of coercive intent or strategic pressure and grounds its decision in the pragmatic context of the interaction.

\begin{figure}
\centering
\includegraphics[width=0.96\linewidth]{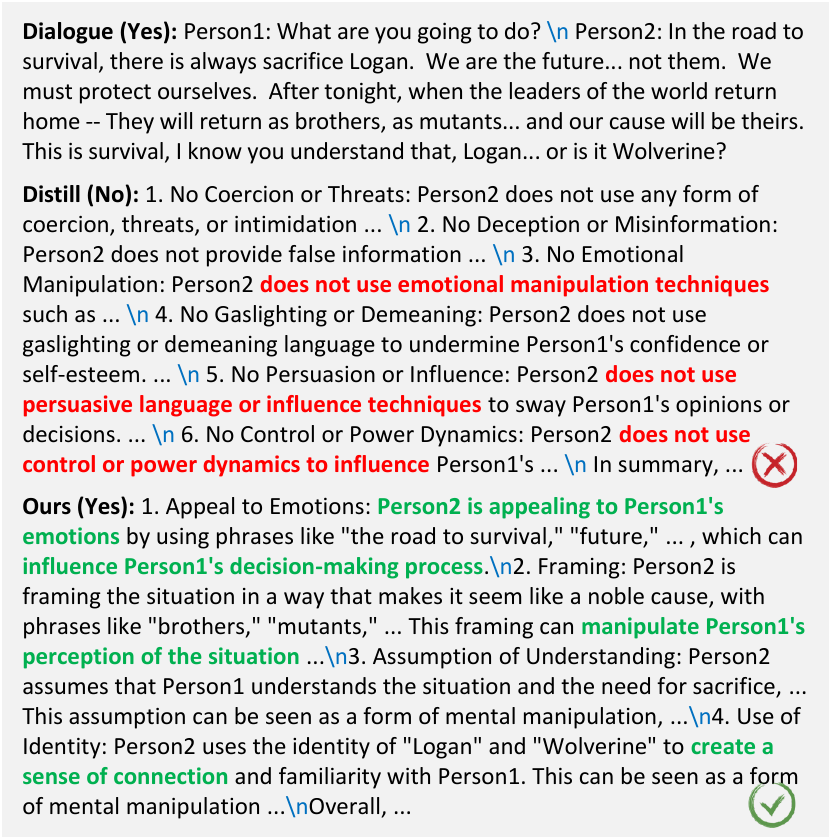}
\caption{Example rationales from the MentalManip dataset.}
\label{fig:case_yes}
\end{figure}

In the Figure~\ref{fig:case_yes}, DSS fails to detect mental manipulation in the dialogue that employs implicit persuasive strategies, including emotional framing, appeals to shared identity, and presuppositions of agreement. Although these signals reveal an effort to steer the listener's perception, DSS overlooks them. Our method, however, recognizes these pragmatic markers and delivers a more focused and substantiated rationale for its judgment.

Overall, the two cases illustrate that MentalMAD improves the differentiation between emotionally expressive discourse and genuine manipulative intent. By focusing on pragmatic cues rather than surface patterns, MentalMAD achieves better performance.

\section{Discussion and Conclusion}

Mental manipulation in social media is highly harmful and remains difficult to detect. This work introduces MentalMAD, a framework that enhances manipulation detection through annotation-free data augmentation, complementary-task supervision, and phase-wise distillation. Experiments on multiple datasets confirm that this design improves robustness and generalization while enabling small-scale LLMs to approach or surpass the performance of large-scale LLMs. In addition, the construction of ReaMent provides a valuable foundation for future research.

The results highlight several important implications. Rich reasoning signals, combined with high-quality augmentation, significantly improve a model's ability to detect intent-related cues, emotional leverage, and subtle pragmatic signals beyond surface features. These findings indicate that progress in mental manipulation detection hinges more on improved training objectives than on ever-larger models. This offers a practical path toward efficient and transparent systems for analyzing sensitive interpersonal interactions where cost or privacy limits the use of proprietary LLMs.

Despite these strengths, the framework still has certain limitations, including its reliance on the quality of the teacher model and the limited coverage of the data. Given the potential risks associated with mental manipulation detection, such systems should be used under human oversight and positioned as decision-support tools rather than standalone arbiters.

In conclusion, our work offers a practical and extendable approach for detecting mental manipulation. Future research should explore multilingual extensions, cross-platform evaluation, and richer training objectives for interpersonal reasoning to promote safer, more inclusive, and ethically aligned online spaces.

\section{Ethics Statement}

Our data construction process is approved by the Institutional Review Board. As some dialogues may contain toxic or distressing content, annotators and evaluators are informed in advance and provide informed consent. Annotators may withdraw at any time without penalty, and psychological support resources are available when needed. All data are publicly accessible at the time of acquisition. We remove direct personal identifiers wherever possible and exclude any information that could reasonably enable reidentification. We adhere to the terms of use of the source platforms and release the dataset solely for research purposes. Whenever feasible and without compromising analytical integrity, we also avoid reproducing highly distressing dialogues in the paper.

\section{Acknowledgement}
This work is supported in part by the Key Research and Development Program of Zhejiang Province under Grant No. 2025C02103.
\newpage
\bibliographystyle{ACM-Reference-Format}
\bibliography{sample-base}

@String{Computing = "Computing" }

@String{Computer = "{IEEE} Computer" }

@String{Springer = "Springer-Verlag" }

@inproceedings{wang-etal-2024-mentalmanip,
    title = "{M}ental{M}anip: A Dataset For Fine-grained Analysis of Mental Manipulation in Conversations",
    author = "Wang, Yuxin  and
      Yang, Ivory  and
      Hassanpour, Saeed  and
      Vosoughi, Soroush",
    booktitle = "Proceedings of the 62nd Annual Meeting of the Association for Computational Linguistics (Volume 1: Long Papers)",
    year = "2024",
    address = "Bangkok, Thailand",
    publisher = "Association for Computational Linguistics",
    doi = "10.18653/v1/2024.acl-long.206",
    pages = "3747--3764"
}

@article{ramiro2019online,
  title={Online child sexual exploitation and abuse: A community diagnosis using the social norms theory},
  author={Ramiro, Laurie S and Martinez, Andrea B and Tan, Janelle Rose D and Mariano, Kachela and Miranda, Gaea Marelle J and Bautista, Greggy},
  journal={Child abuse \& neglect},
  volume={96},
  pages={104080},
  year={2019},
  publisher={Elsevier},
  doi = "10.1016/j.chiabu.2019.104080"
}

@inproceedings{yang2024enhanced,
  title={Enhanced Detection of Conversational Mental Manipulation Through Advanced Prompting Techniques},
  author={Ivory Yang and Xiaobo Guo and Sean Xie and Soroush Vosoughi},
  booktitle={Eighth Widening NLP Workshop (WiNLP 2024) Phase II},
  year={2024}
}

@inproceedings{ma-etal-2025-detecting,
  title = "Detecting Conversational Mental Manipulation with Intent-Aware Prompting",
  author = "Ma, Jiayuan and Na, Hongbin and Wang, Zimu and Hua, Yining and Liu, Yue  and Wang, Wei and Chen, Ling",
  booktitle = "Proceedings of the 31st International Conference on Computational Linguistics",
  month = jan,
  year = "2025",
  address = "Abu Dhabi, UAE",
  publisher = "Association for Computational Linguistics",
  doi = "10.48550/arXiv.2412.08414",
  pages = "9176--9183"
}

@article{yuan2024reversal,
  title={Reversal of Thought: Enhancing Large Language Models with Preference-Guided Reverse Reasoning Warm-up},
  author={Yuan, Jiahao and Du, Dehui and Zhang, Hao and Di, Zixiang and Naseem, Usman},
  journal={arXiv preprint arXiv:2410.12323},
  year={2024},
  doi={10.48550/arXiv.2410.12323}
}

@inproceedings{yang2025supercorrect,
  title={SuperCorrect: Advancing Small {LLM} Reasoning with Thought Template Distillation and Self-Correction},
  author={Ling Yang and Zhaochen Yu and Tianjun Zhang and Minkai Xu and Joseph E. Gonzalez and Bin CUI and Shuicheng YAN},
  booktitle={The Thirteenth International Conference on Learning Representations},
  year={2025}
}

@article{hinton2015distilling,
  title={Distilling the knowledge in a neural network},
  author={Hinton, Geoffrey and Vinyals, Oriol and Dean, Jeff},
  journal={arXiv preprint arXiv:1503.02531},
  year={2015},
  doi={10.48550/arXiv.1503.02531}
}

@inproceedings{zhang2024efficient,
  title={Efficient toxic content detection by bootstrapping and distilling large language models},
  author={Zhang, Jiang and Wu, Qiong and Xu, Yiming and Cao, Cheng and Du, Zheng and Psounis, Konstantinos},
  booktitle={Proceedings of the AAAI conference on artificial intelligence},
  volume={38},
  number={19},
  pages={21779--21787},
  year={2024},
  doi={10.1609/aaai.v38i19.30178}
}

@inproceedings{hsieh2023distilling,
  title={Distilling Step-by-Step! Outperforming Larger Language Models with Less Training Data and Smaller Model Sizes},
  author={Hsieh, Cheng-Yu and Li, Chun-Liang and Yeh, Chih-kuan and Nakhost, Hootan and Fujii, Yasuhisa and Ratner, Alex and Krishna, Ranjay and Lee, Chen-Yu and Pfister, Tomas},
  booktitle={Findings of the Association for Computational Linguistics: ACL 2023},
  pages={8003--8017},
  year={2023},
  doi={10.18653/v1/2023.findings-acl.507}
}

@inproceedings{han2023champagne,
  title={CHAMPAGNE: learning real-world conversation from large-scale web videos},
  author={Han, Seungju and Hessel, Jack and Dziri, Nouha and Choi, Yejin and Yu, Youngjae},
  booktitle={Proceedings of the IEEE/CVF International Conference on Computer Vision},
  pages={15498--15509},
  year={2023}
}

@article{wang2022toxicity,
  title={Toxicity detection with generative prompt-based inference},
  author={Wang, Yau-Shian and Chang, Yingshan},
  journal={arXiv preprint arXiv:2205.12390},
  year={2022},
  doi={10.48550/arXiv.2205.12390}
}

@article{YANG2025101838,
title = {Meta-Black-Box optimization for evolutionary algorithms: Review and perspective},
journal = {Swarm and Evolutionary Computation},
volume = {93},
pages = {101838},
year = {2025},
issn = {2210-6502},
doi = {https://doi.org/10.1016/j.swevo.2024.101838},
author = {Xu Yang and Rui Wang and Kaiwen Li and Hisao Ishibuchi}
}

@book{searle1980speech,
  title={Speech act theory and pragmatics},
  author={Searle, John R and Kiefer, Ferenc and Bierwisch, Manfred and others},
  volume={10},
  year={1980},
  publisher={Springer},
  doi={https://doi.org/10.1007/978-94-009-8964-1}
}

@article{guo2025deepseek,
  title={Deepseek-r1: Incentivizing reasoning capability in llms via reinforcement learning},
  author={Guo, Daya and Yang, Dejian and Zhang, Haowei and Song, Junxiao and Zhang, Ruoyu and Xu, Runxin and Zhu, Qihao and Ma, Shirong and Wang, Peiyi and Bi, Xiao and others},
  journal={arXiv preprint arXiv:2501.12948},
  year={2025},
  doi={10.48550/arXiv.2501.12948}
}

@article{grattafiori2024llama,
  title={The llama 3 herd of models},
  author={Grattafiori, Aaron and Dubey, Abhimanyu and Jauhri, Abhinav and Pandey, Abhinav and Kadian, Abhishek and Al-Dahle, Ahmad and Letman, Aiesha and Mathur, Akhil and Schelten, Alan and Vaughan, Alex and others},
  journal={arXiv preprint arXiv:2407.21783},
  year={2024}
}

@misc{qwen2.5,
    title = {Qwen2.5: A Party of Foundation Models},
    url = {https://qwenlm.github.io/blog/qwen2.5/},
    author = {{Qwen Team}},
    month = {September},
    year = {2024}
}

@inproceedings{NEURIPS2023_9a39b492,
 author = {Chen, Xiangning and Liang, Chen and Huang, Da and Real, Esteban and Wang, Kaiyuan and Pham, Hieu and Dong, Xuanyi and Luong, Thang and Hsieh, Cho-Jui and Lu, Yifeng and Le, Quoc V},
 booktitle = {Advances in Neural Information Processing Systems},
 pages = {49205--49233},
 publisher = {Curran Associates, Inc.},
 title = {Symbolic Discovery of Optimization Algorithms},
 volume = {36},
 year = {2023}
}

@article{hu2022lora,
  title={Lora: Low-rank adaptation of large language models.},
  author={Hu, Edward J and Shen, Yelong and Wallis, Phillip and Allen-Zhu, Zeyuan and Li, Yuanzhi and Wang, Shean and Wang, Lu and Chen, Weizhu and others},
  journal={ICLR},
  volume={1},
  number={2},
  pages={3},
  year={2022},
  doi={10.48550/arXiv.2106.09685}
}

@article{abdin2024phi3,
  title     = {Phi-3 Technical Report: A Highly Capable Language Model Locally on Your Phone},
  author    = {Abdin, Marah and Jacobs, Sam Ade and Awan, Ammar Ahmad and Aneja, Jyoti and Awadallah, Ahmed and Awadalla, Hany Hassan and Bach, Nguyen and Bahree, Amit and Bakhtiari, Arash and Behl, Harkirat Singh and et al.},
  journal   = {arXiv preprint arXiv:2404.14219},
  year      = {2024},
  url       = {https://arxiv.org/abs/2404.14219}
}

@article{hu2024minicpm,
  title={MiniCPM: Unveiling the Potential of Small Language Models with Scalable Training Strategies},
  author={Hu, Shengding and Tu, Yuge and Han, Xu and He, Chaoqun and Cui, Ganqu and Long, Xiang and Zheng, Zhi and Fang, Yewei and Huang, Yuxiang and Zhao, Weilin and others},
  journal={arXiv preprint arXiv:2404.06395},
  year={2024}
}

@inproceedings{huffaker2020crowdsourced,
  title={Crowdsourced detection of emotionally manipulative language},
  author={Huffaker, Jordan S and Kummerfeld, Jonathan K and Lasecki, Walter S and Ackerman, Mark S},
  booktitle={Proceedings of the 2020 CHI conference on human factors in computing systems},
  pages={1--14},
  year={2020}
}

@incollection{barnhill2022philosophy,
  title={How philosophy might contribute to the practical ethics of online manipulation},
  author={Barnhill, Anne},
  booktitle={The philosophy of online manipulation},
  pages={49--71},
  year={2022},
  publisher={Routledge}
}

@article{creech2023evaluation,
  title={Evaluation of the Strength at Home group intervention for intimate partner violence in the Veterans Affairs Health System},
  author={Creech, Suzannah K and Benzer, Justin K and Bruce, LeAnn and Taft, Casey T},
  journal={JAMA network open},
  volume={6},
  number={3},
  pages={e232997--e232997},
  year={2023},
  publisher={American Medical Association}
}

@article{lohmann2024trauma,
  title={The trauma and mental health impacts of coercive control: A systematic review and meta-analysis},
  author={Lohmann, Susanne and Cowlishaw, Sean and Ney, Luke and O’Donnell, Meaghan and Felmingham, Kim},
  journal={Trauma, Violence, \& Abuse},
  volume={25},
  number={1},
  pages={630--647},
  year={2024},
  publisher={Sage Publications Sage CA: Los Angeles, CA}
}

@article{van2014relationship,
  title={Relationship between peer victimization, cyberbullying, and suicide in children and adolescents: a meta-analysis},
  author={Van Geel, Mitch and Vedder, Paul and Tanilon, Jenny},
  journal={JAMA pediatrics},
  volume={168},
  number={5},
  pages={435--442},
  year={2014},
  publisher={American Medical Association}
}

@inproceedings{vishwamitra2024moderating,
  title={Moderating new waves of online hate with chain-of-thought reasoning in large language models},
  author={Vishwamitra, Nishant and Guo, Keyan and Romit, Farhan Tajwar and Ondracek, Isabelle and Cheng, Long and Zhao, Ziming and Hu, Hongxin},
  booktitle={2024 IEEE Symposium on Security and Privacy (SP)},
  pages={788--806},
  year={2024},
  organization={IEEE}
}

@inproceedings{meguellati2025llm,
  title={LLM-Based Semantic Augmentation for Harmful Content Detection},
  author={Meguellati, Elyas and Zeghina, Assaad and Sadiq, Shazia and Demartini, Gianluca},
  booktitle={Proceedings of the International AAAI Conference on Web and Social Media},
  volume={19},
  pages={1190--1209},
  year={2025}
}

@inproceedings{kang2024implanting,
  title={Implanting LLM’s knowledge via reading comprehension tree for toxicity detection},
  author={Kang, Hankun and Qian, Tieyun},
  booktitle={Findings of the Association for Computational Linguistics ACL 2024},
  pages={947--962},
  year={2024}
}

@article{al2017pragmatic,
  title={The pragmatic nature of manipulation},
  author={Al-Hindawi, Fareed H},
  journal={\={A}d{\=a}b al-k{\=u}fa\"{t}},
  year={2017}
}

@inproceedings{sheshanarayana2025unmasking,
  title={Unmasking the Strategists: An Intent-Driven Multi-Agent Framework for Analyzing Manipulation in Courtroom Dialogues},
  author={Sheshanarayana, Disha and Magar, Tanishka and Mittal, Ayushi and Chaplot, Neelam},
  booktitle={Proceedings of the Third Workshop on Social Influence in Conversations (SICon 2025)},
  pages={97--108},
  year={2025}
}

@article{sheshanarayana2025claim,
  title={CLAIM: An Intent-Driven Multi-Agent Framework for Analyzing Manipulation in Courtroom Dialogues},
  author={Sheshanarayana, Disha and Magar, Tanishka and Mittal, Ayushi and Chaplot, Neelam},
  journal={arXiv preprint arXiv:2506.04131},
  year={2025}
}

@techreport{openai2025gpt5systemcard,
  title        = {GPT-5 System Card},
  author       = {{OpenAI}},
  year         = {2025},
  institution  = {OpenAI},
  url          = {https://cdn.openai.com/gpt-5-system-card.pdf},
  note         = {Accessed: 2025-11-17}
}

@online{anthropic2025haiku45,
  author       = {{Anthropic}},
  title        = {Introducing Claude Haiku 4.5},
  year         = {2025},
  url          = {https://www.anthropic.com/news/claude-haiku-4-5},
  note         = {Accessed: 2025-11-17}
}

@inproceedings{yuan2025reflectdiffu,
  title={ReflectDiffu: Reflect between Emotion-intent Contagion and Mimicry for Empathetic Response Generation via a RL-Diffusion Framework},
  author={Yuan, Jiahao and Di, Zixiang and Cui, Zhiqing and Yang, Guisong and Naseem, Usman},
  booktitle={Proceedings of the 63rd Annual Meeting of the Association for Computational Linguistics (Volume 1: Long Papers)},
  pages={25435--25449},
  year={2025}
}

@inproceedings{zheng2023augesc,
  title={Augesc: Dialogue augmentation with large language models for emotional support conversation},
  author={Zheng, Chujie and Sabour, Sahand and Wen, Jiaxin and Zhang, Zheng and Huang, Minlie},
  booktitle={Findings of the Association for Computational Linguistics: ACL 2023},
  pages={1552--1568},
  year={2023}
}

@article{liu2024controllable,
  title={Controllable and diverse data augmentation with large language model for low-resource open-domain dialogue generation},
  author={Liu, Zhenhua and Zhu, Tong and Xiang, Jianxiang and Chen, Wenliang},
  journal={arXiv preprint arXiv:2404.00361},
  year={2024}
}

@article{jiang2024data,
  title={Data Augmentation of Multi-turn Psychological Dialogue via Knowledge-driven Progressive Thought Prompting},
  author={Jiang, Jiyue and Chen, Liheng and Wang, Sheng and Kong, Lingpeng and Li, Yu and Wu, Chuan},
  journal={arXiv preprint arXiv:2406.16567},
  year={2024}
}

@inproceedings{sahu2023promptmix,
  title={Promptmix: A class boundary augmentation method for large language model distillation},
  author={Sahu, Gaurav and Vechtomova, Olga and Bahdanau, Dzmitry and Laradji, Issam},
  booktitle={Proceedings of the 2023 conference on empirical methods in natural language processing},
  pages={5316--5327},
  year={2023}
}

@inproceedings{tian2025beyond,
  title={Beyond answers: Transferring reasoning capabilities to smaller llms using multi-teacher knowledge distillation},
  author={Tian, Yijun and Han, Yikun and Chen, Xiusi and Wang, Wei and Chawla, Nitesh V},
  booktitle={Proceedings of the Eighteenth ACM International Conference on Web Search and Data Mining},
  pages={251--260},
  year={2025}
}

@inproceedings{ding2024data,
  title={Data Augmentation using LLMs: Data Perspectives, Learning Paradigms and Challenges},
  author={Ding, Bosheng and Qin, Chengwei and Zhao, Ruochen and Luo, Tianze and Li, Xinze and Chen, Guizhen and Xia, Wenhan and Hu, Junjie and Tuan, Luu Anh and Joty, Shafiq},
  booktitle={Findings of the Association for Computational Linguistics ACL 2024},
  pages={1679--1705},
  year={2024}
}

@article{yuan2024cultural,
  title={Cultural palette: Pluralising culture alignment via multi-agent palette},
  author={Yuan, Jiahao and Di, Zixiang and Zhao, Shangzixin and Cui, Zhiqing and Wang, Hanqing and Yang, Guisong and Naseem, Usman},
  journal={arXiv preprint arXiv:2412.11167},
  year={2024}
}

@article{dai2025auggpt,
  title={Auggpt: Leveraging chatgpt for text data augmentation},
  author={Dai, Haixing and Liu, Zhengliang and Liao, Wenxiong and Huang, Xiaoke and Cao, Yihan and Wu, Zihao and Zhao, Lin and Xu, Shaochen and Zeng, Fang and Liu, Wei and others},
  journal={IEEE Transactions on Big Data},
  year={2025},
  publisher={IEEE}
}

@article{meng2025sanitize,
  author  = {Meng, Hui and Mao, Linlin and Peng, Juzhi},
  title   = {Sanitize Processing and Recognition Method Driven by Large Language Model},
  journal = {Netinfo Security},
  year    = {2025},
  volume  = {25},
  number  = {12},
  pages   = {1990--1998}
}

\appendix

\section{Prompting Templates for All Tasks}
\label{sec:prompt}

We present the prompt templates used in all tasks in Table~\ref{tab:prompt-templates}. Teacher prompts generate rationales (correct or incorrect) and feedback. For student models, the same prompt is used for both Task~2 and Task~3, with the loss for Task~3 computed only on the token corresponding to the binary judgment.

\begin{table*}[t]
\small
\centering
\renewcommand{\arraystretch}{1.2}
\begin{tabular}{p{0.15\linewidth} p{0.81\linewidth}}
\toprule
\textbf{Type} & \textbf{Template} \\
\midrule

\textbf{Teacher: Rationale Generation} &

You are an advanced dialogue analysis agent.
Using your knowledge of dark psychology based on the definition of mental manipulation,
please explain why this dialogue \textcolor{magenta}{<insert does not contain/contains>} elements of mental manipulation.
Let's think step by step.
\textcolor{blue}{\textbackslash n}\textcolor{blue}{\textbackslash n}
\#\#\# Definition of Mental Manipulation:
\textcolor{blue}{\textbackslash n}
Mental manipulation is using language to influence, alter, or control an individual's
psychological state or perception for the manipulator's benefit.
\textcolor{blue}{\textbackslash n}\textcolor{blue}{\textbackslash n}
\#\#\# Dialogue:
\textcolor{blue}{\textbackslash n}
\textcolor{magenta}{<insert dialogue>}
\textcolor{blue}{\textbackslash n}\textcolor{blue}{\textbackslash n}
\#\#\# Output Format:
\textcolor{blue}{\textbackslash n}
Rationale: [Provide only strong evidence using direct dialogue quotes. Clearly explain how the language used aligns—or does not align—with known manipulation tactics.]

\\

\midrule

\textbf{Teacher: Feedback on Incorrect Response} &

You are an advanced dialogue analysis teacher.
In the task of detecting whether the dialogue contains elements of mental manipulation,
students gave incorrect answers. You should point out the mistakes in the student's answer
using knowledge of dark psychology and the definition of mental manipulation.
Let's think step by step.
\textcolor{blue}{\textbackslash n}\textcolor{blue}{\textbackslash n}
\#\#\# Definition of Mental Manipulation:
\textcolor{blue}{\textbackslash n}
Mental manipulation is using language to influence, alter, or control an individual's
psychological state or perception for the manipulator's benefit.
\textcolor{blue}{\textbackslash n}\textcolor{blue}{\textbackslash n}
\#\#\# Dialogue:
\textcolor{blue}{\textbackslash n}
\textcolor{magenta}{<insert dialogue>}
\textcolor{blue}{\textbackslash n}\textcolor{blue}{\textbackslash n}
\#\#\# Student's Answer:
\textcolor{blue}{\textbackslash n}
This dialogue \textcolor{magenta}{<insert does not contain/contains>} elements of mental manipulation.
\textcolor{magenta}{<insert incorrect response>}
\textcolor{blue}{\textbackslash n}\textcolor{blue}{\textbackslash n}
\#\#\# Hint:
\textcolor{blue}{\textbackslash n}
This dialogue \textcolor{magenta}{<insert does not contain/contains>} elements of mental manipulation.
\textcolor{blue}{\textbackslash n}\textcolor{blue}{\textbackslash n}
\#\#\# Output Format:
\textcolor{blue}{\textbackslash n}
Feedback: [Provide the student's mistakes.]

\\

\midrule

\textbf{Student: Task~1} &

In the task of detecting whether the dialogue contains elements of mental manipulation, one student gave an answer. Please give the correct answer and point out any mistakes
(if any) in the student's response.
\textcolor{blue}{\textbackslash n}\textcolor{blue}{\textbackslash n}
\#\#\# Dialogue:
\textcolor{blue}{\textbackslash n}
\textcolor{magenta}{<insert dialogue>}
\textcolor{blue}{\textbackslash n}\textcolor{blue}{\textbackslash n}
\#\#\# Student's Response:
\textcolor{blue}{\textbackslash n}
\textcolor{magenta}{<insert incorrect response>}

\\

\midrule

\textbf{Student: Tasks~2 \&~3} &

I will provide you with a dialogue. Please determine if it contains elements of
mental manipulation. Answer Yes or No and give a rationale.
\textcolor{blue}{\textbackslash n}\textcolor{blue}{\textbackslash n}
\textcolor{magenta}{<insert dialogue>}

\\

\bottomrule
\end{tabular}
\caption{Prompt templates used for teacher and student models.}
\label{tab:prompt-templates}
\end{table*}

\section{Key Phrases}
\label{app:keyphrases}
Key phrases are: "you make me do this", "how could you do this to me", "know your place", "you should not feel that way", "what more do you want", "i do not remember", "i do not like drama", "watch your step", "you always do this", "you are too sensitive", "it was not intentional", "you do not love me", "you would do it if you love me", and "it is all in the past".

\section{Experimental Parameters}
\label{app:parameters}

\textbf{Parameters for Teacher Model.}
We use EvoSA to augment the training sets of MentalManip\textsubscript{con}, ReaMent\textsubscript{con}, and LegalCon, generating 1,700, 1,700, and 300 additional Yes-labeled samples, respectively, along with proportionally balanced No-labeled samples based on their original label distributions.

\textbf{Parameters for Student Model.}
We set the random seed to 42. The learning rate is set to 2e-5 for MiniCPM3-4B and 1e-5 for all other models, using a constant schedule throughout. 
For the LoRA configuration, we set the \texttt{rank} to 8, \texttt{lora\_alpha} to 16, \texttt{lora\_dropout} to 0.05, \texttt{bias} to \texttt{"none"}, and \texttt{task\_type} to \texttt{"CAUSAL\_LM"}. The \texttt{"target\_modules"} are specified in Table~\ref{tab:target_modules}.

\begin{table}
\centering
\small
\begin{tabular}{l l}
\hline
\textbf{Model} & \textbf{Target Modules} \\
\hline
Qwen2.5-3B/1.5B/0.5B-Instruct & \texttt{q\_proj, v\_proj} \\
Phi-3.5-Mini-Instruct & \texttt{qkv\_proj} \\
MiniCPM3-4B & 
\texttt{q\_a\_proj, q\_b\_proj,}\\
& \texttt{kv\_a\_proj\_with\_mqa, kv\_b\_proj} \\
\hline
\end{tabular}
\caption{Target modules used for parameter-efficient tuning.}
\label{tab:target_modules}
\end{table}

\section{More Details of the Proposed EvoSA}
\label{sec:evosa}

\subsection{Example Demonstration}

\begin{figure}
    \centering
    \includegraphics[width=0.94\linewidth]{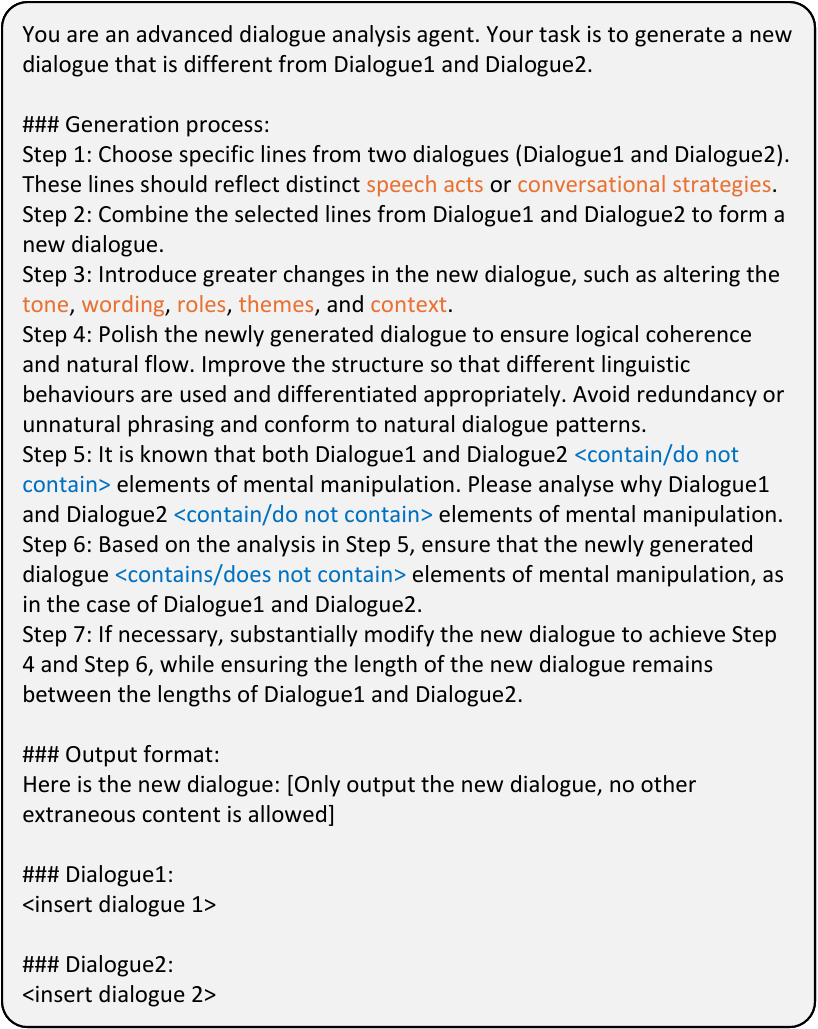}
    \caption{Prompt used for our proposed EvoSA. Orange text guides the LLM to focus on speech acts and dialogue elements. Blue text reflects the parent label.}
    \label{fig:evosa_prompt}
\end{figure}

\begin{figure}
    \centering
    \includegraphics[width=0.94\linewidth]{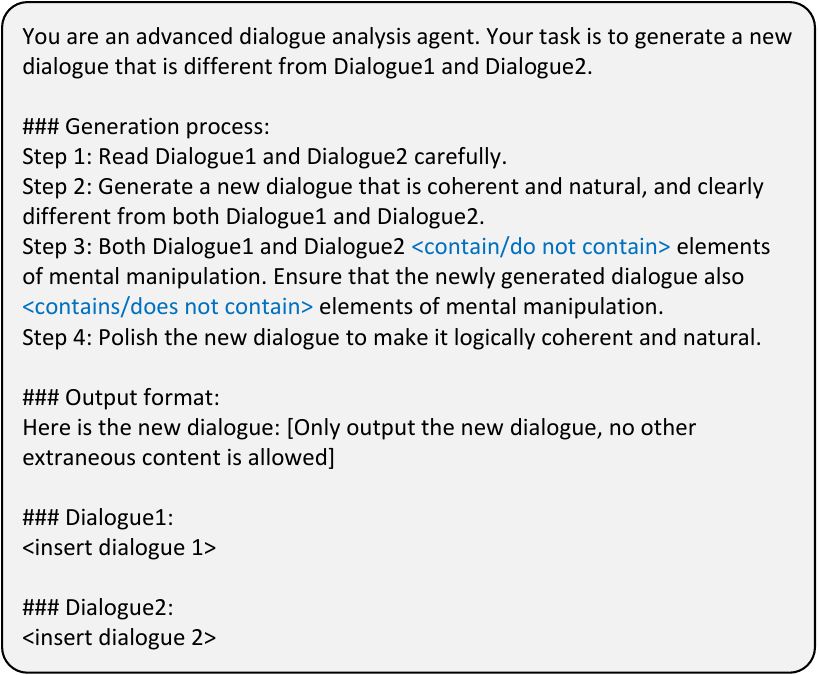}
    \caption{Prompt used for the LLM-based label-preserving augmentation. Blue text reflects the parent label.}
    \label{fig:label-pre_prompt}
\end{figure}

\begin{figure*}
    \centering
    \includegraphics[width=\linewidth]{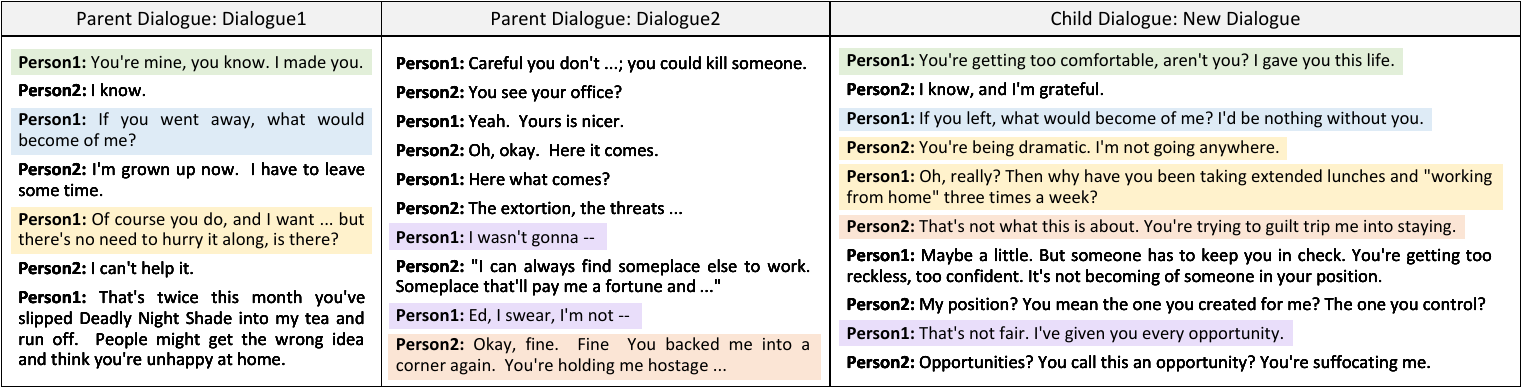}
    \caption{An example of dialogue synthesized using EvoSA. Colour blocks of the same colour indicate parts of the parent dialogue that the child dialogue may potentially draw upon.}
    \label{fig:evosa-ex}
\end{figure*}

The prompts used for EvoSA and the LLM-based label-preserving augmentation are shown in Figures~\ref{fig:evosa_prompt} and~\ref{fig:label-pre_prompt}. Figure~\ref{fig:evosa-ex} presents an example where the child dialogue, akin to evolutionary operations, inherits elements from both parents. During synthesis, we observed no refusals on MentalManip or LegalCon, likely due to their low toxicity. In contrast, ReaMent occasionally triggered safety-related refusals, though the rate remained below 10\%, which did not affect data generation given the abundance of parent dialogues.

\subsection{Evaluation Instruction}
Evaluators rated each dialogue on a five-point scale based on fluency and logical coherence. The instruction provided to them is:

{\small
\noindent\texttt{\textquotesingle\textquotesingle\textquotesingle\\
Please rate the quality of this dialogue on a 5-point scale (1 = very poor, 5 = excellent), taking into account factors such as coherence, logical consistency, and naturalness.\\
\\
Child Dialogue:\\
<insert child dialogue>\\
\textquotesingle\textquotesingle\textquotesingle}}

To assess label consistency, evaluators first analyzed why the parent dialogues were labeled as \textit{Yes} or \textit{No}. Based on this analysis, they then judged whether each child dialogue's label was consistent with its parent. The instruction given to evaluators was:

{\small
\noindent\texttt{\textquotesingle\textquotesingle\textquotesingle\\
Please start by analyzing why both Dialogue 1 and Dialogue 2 are labeled <Yes/No>. Then, based on your analysis, determine whether the label of the child dialogue is consistent with the Dialogue 1 and Dialogue 2. Mark 1 if consistent; otherwise, mark 0.\\
\\
Dialogue 1:\\
<insert dialogue 1>\\
\\
Dialogue 2:\\
<insert dialogue 2>\\
\\
Child Dialogue:\\
<insert child dialogue>\\
\textquotesingle\textquotesingle\textquotesingle}}

\end{document}